\documentclass[10pt,twocolumn,letterpaper]{article}

\usepackage{cvpr}              

\usepackage{graphicx}
\usepackage{amsmath}
\usepackage{amssymb}
\usepackage{booktabs}
\usepackage{multirow}
\usepackage[table,xcdraw]{xcolor}
\usepackage{makecell}
\usepackage{subfloat}
\usepackage{balance} 
\usepackage{soul}
\usepackage[accsupp]{axessibility}
\usepackage[autostyle]{csquotes}
\usepackage[pagebackref,breaklinks,colorlinks]{hyperref}

\usepackage[capitalize]{cleveref}
\crefname{section}{Sec.}{Secs.}
\Crefname{section}{Section}{Sections}
\Crefname{table}{Table}{Tables}
\crefname{table}{Tab.}{Tabs.}

\def\cvprPaperID{1142} 
\def\confName{CVPR}
\def\confYear{2022}

\begin{document}

\title{Fine-Grained Predicates Learning for Scene Graph Generation}


\author{Xinyu Lyu\textsuperscript{1} 
	\and Lianli Gao\textsuperscript{1} \thanks{Corresponding author.}
	\and Yuyu Guo\textsuperscript{1} 
	\and Zhou Zhao\textsuperscript{2} 
	\and Hao Huang\textsuperscript{3} 
	\and Heng Tao Shen\textsuperscript{1} 
	\and Jingkuan Song\textsuperscript{1}  \\ \textsuperscript{1}Center for Future Media \& School of Computer Science and Engineering, \\ University of Electronic Science and Technology of China, China \\ \textsuperscript{2}Zhejiang University, China \\ \textsuperscript{3}Kuaishou, China \\
}


\maketitle

\begin{abstract}
The performance of current Scene Graph Generation models is severely hampered by some hard-to-distinguish predicates, e.g., ``woman-\underline{on}/\underline{standing on}/\underline{walking on}-beach'' or ``woman-\underline{near}/\underline{looking at}/\underline{in front of}-child''. 
While general SGG models are prone to predict head predicates and existing re-balancing strategies prefer tail categories, none of them can appropriately handle these hard-to-distinguish predicates. 
To tackle this issue, inspired by fine-grained image classification, which focuses on differentiating among hard-to-distinguish object classes, we propose a method named \textbf{Fine-Grained Predicates Learning (FGPL)} which aims at differentiating among hard-to-distinguish predicates for Scene Graph Generation task. Specifically, we first introduce a \noindent\textbf{Predicate Lattice} that helps SGG models to figure out fine-grained predicate pairs. Then, utilizing the Predicate Lattice, we propose a \noindent\textbf{Category Discriminating Loss} and an \noindent\textbf{Entity Discriminating Loss}, which both contribute to distinguishing fine-grained predicates while maintaining learned discriminatory power over recognizable ones. The proposed model-agnostic strategy \textbf{significantly} boosts the performances of three benchmark models (Transformer, VCTree, and Motif) by \noindent\textbf{22.8$\%$, 24.1$\%$ and 21.7$\%$ of Mean Recall (mR@100)} on the Predicate Classification sub-task, respectively. Our model also outperforms state-of-the-art methods by a large margin (i.e., \noindent\textbf{6.1$\%$, 4.6$\%$, and 3.2$\%$ of Mean Recall (mR@100)}) on the Visual Genome dataset.

\end{abstract}

\begin{figure}[htbp]
\begin{center}
\includegraphics[width=0.47\textwidth]{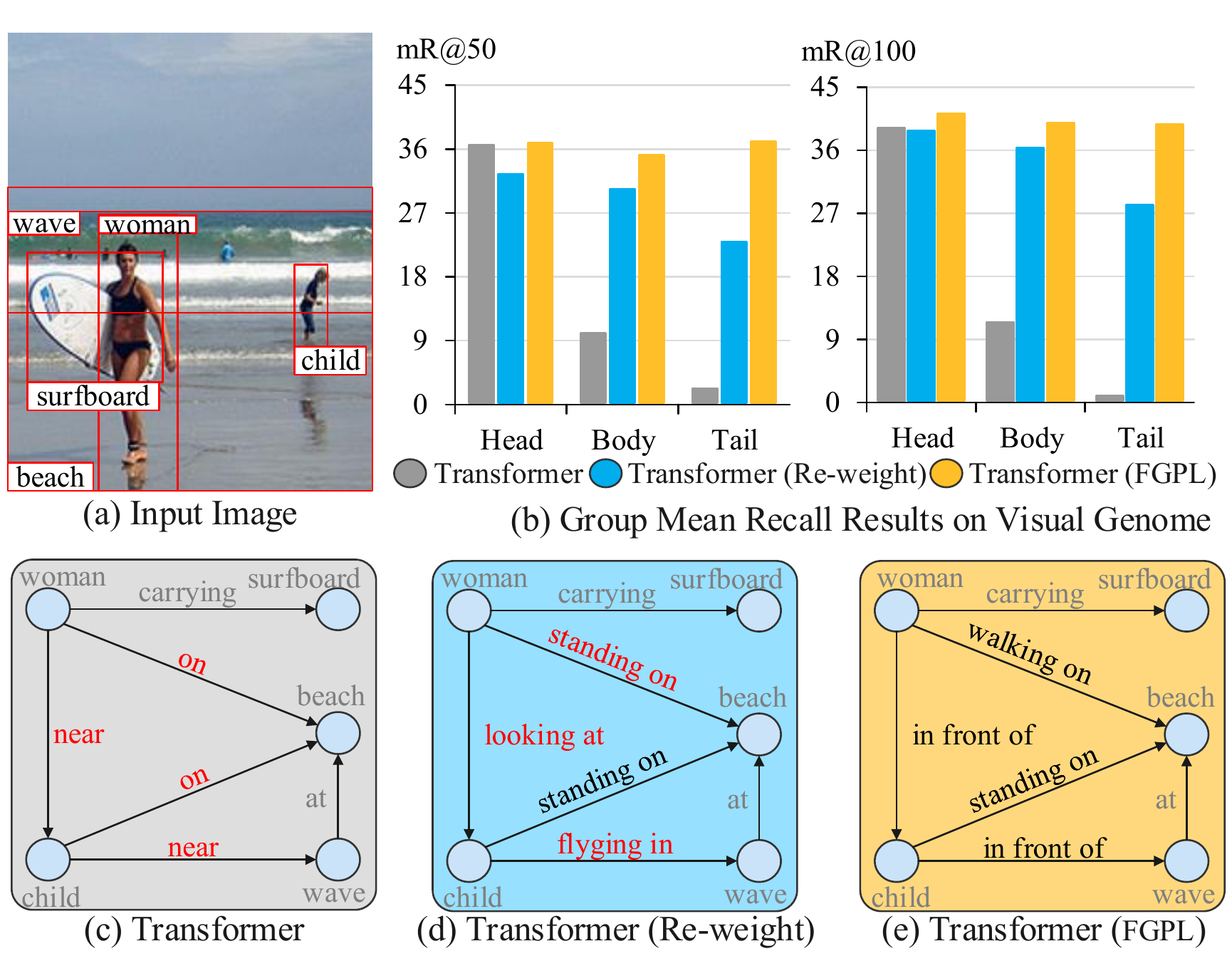}
\vspace{-1.0em}
\caption{\textbf{The illustration of handling hard-to-distinguish predicates for SSG models.} (b) Transformer (FGPL) outperforms both Transformer and Transformer (Re-weight) on Group Mean Recall. (c) Transformer~\cite{networks:transformer,ssg:benchmark} is prone to predict head predicates. (d) Transformer (Re-weight) prefers tail categories. (e) Transformer (FGPL) can appropriately handle hard-to-distinguish predicates, \eg, ``woman-on/standing on/\textbf{walking on}-beach'' or ``woman-near/looking at/\textbf{in front of}-child''.}
\label{fig:abstract}
\vspace{-1.5em}
\end{center}
\end{figure}

\section{Introduction}
\label{sec:intro}
Scene graph generation plays a vital role in visual understanding, which intends to detect instances together with their relationships. By ultimately representing image contents in a graph structure, scene graph generation serves as a powerful means to bridge the gap between visual scenes and human languages, benefiting several visual-understanding tasks, such as image retrieval~\cite{image_retrival,DBLP:conf/mm/ZengGLJS21}, image captioning~\cite{cap1,cap2}, and visual question answering~\cite{ssg:vctree,qa1,qa3,qa4,DBLP:conf/aaai/LiSGLH0G19,DBLP:conf/ijcai/SongZGS18,wu2021star}. 

\begin{figure}[t]
\begin{center}
\includegraphics[width=0.4\textwidth]{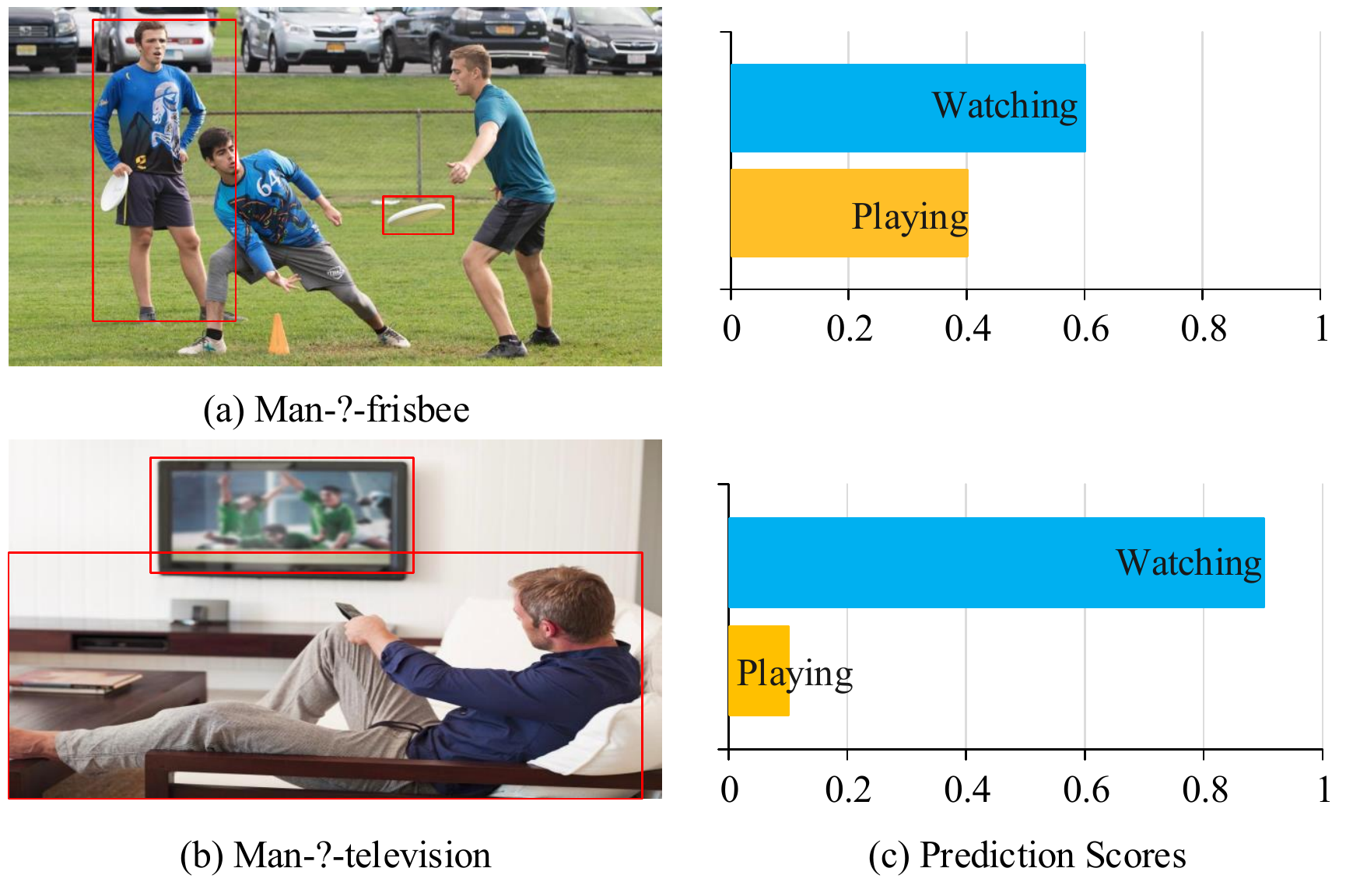}
\vspace{-0.5em}
\caption{\textbf{The illustration of predicate correlations concerning contexts.} The predicate correlation between ``watching'' and ``playing'' varies with contexts. Especially, ``watching/playing'' is weakly correlated or distinguishable in (b), while they are strongly correlated or hard-to-distinguish in (a).}
\label{fig:problem}
\vspace{-2.0em}
\end{center}
\end{figure}

Prior works~\cite{imp,motif,ssg:vctree,gps,vrr,energy,DBLP:conf/mm/GuoSGS20} have devoted great efforts to exploring representation learning for scene graph generation, but the biased prediction is still challenging because of the long-tailed distribution of predicates in SGG datasets. Trained with severely skewed class distributions, general SGG models are prone to predict head predicates, as results of Transformer~\cite{networks:transformer,ssg:benchmark} shown in Fig.~\ref{fig:abstract}(c). Recent works~\cite{cogtree,prior:iccv,bgnn,sample} have exploited re-balancing methods to solve the biased prediction problem for scene graph generation, making predicates distribution balanced or the learning process smooth. As demonstrated in Fig.~\ref{fig:abstract}(b), Transformer (Re-weight) achieves a more balanced performance than Transformer. However, relying on the class distribution, existing re-balancing strategies prefer predicates from tail categories while being hampered by some hard-to-distinguish predicates. For instance, as shown in Fig.~\ref{fig:abstract}(d), Transformer (Re-weight) misclassifies \textit{``woman-\underline{in front of}-child''} as \textit{``woman-\underline{looking at}-child''} in terms of visual correlations between ``in front of'' and ``looking at''. 
The origin of the issue lies in the fact that differentiating hard-to-distinguish predicates requires exploring their correlations. Underestimating correlations among predicates, existing methods~\cite{cogtree,pcpl} cannot choose hard-to-distinguish ones for sufficient punishment. To acquire complete predicate correlations, we consider contextual information since correlations between a pair of predicates may dramatically vary with contexts as stated in~\cite{nlp}. Particularly, contexts are regarded as visual or semantic information of predicates' objects and subjects in scene graph generation. Take predicate correlations analysis between ``watching'' and ``playing'' as an example. ``Watching/playing'' is weakly correlated or distinguishable in Fig.~\ref{fig:problem}(b), while they are strongly correlated or hard-to-distinguish in Fig.~\ref{fig:problem}(a).

Inspired by the above observations, we propose a Fine-Grained Predicates Learning (FGPL) framework by thoroughly exploiting predicate correlations. We first introduce a Predicate Lattice to help understand ubiquitous predicate correlations concerning all scenarios in the SGG dataset. With the Predicate Lattice, we devise a Category Discriminating Loss (CDL) and an Entity Discriminating Loss (EDL), which both discriminate hard-to-distinguish predicates while maintaining learned discriminatory power over recognizable ones. In particular, Category Discriminating Loss (CDL) attempts to figure out and differentiate hard-to-distinguish predicates. Furthermore, as predicates' correlation varies with contexts of entities, Entity Discriminating Loss (EDL) adaptively adjusts the discriminating process according to predictions of entities. Using CDL and EDL, our method can determine whether predicate pairs are hard-to-distinguish or not during training, which guarantees a more balanced learning process among different categories than previous methods~\cite{cogtree,pcpl,prior:iccv,bgnn,sample}.

\noindent\textbf{Contribution}: Our main contributions are summarized as follows: 1). We propose a novel plug-and-play Fine-Grained Predicates Learning (FGPL) framework to differentiate hard-to-distinguish predicates for scene graph generation. 2). We devise a Predicate Lattice to obtain complete predicate correlations between each predicate pair concerning context information. Category Discriminating Loss (CDL) aims at figuring out and differentiating hard-to-distinguish predicates. Moreover, Entity Discriminating Loss (EDL) adaptively adjusts the discriminating process according to predictions of entities. 3). Our FGPL greatly boosts performances of three benchmark models (Transformer, VCTree, and Motif) by 22.8$\%$, 24.1$\%$, and 21.7$\%$ of Mean Recall (mR@100) on Predicate Classification sub-task and achieves superior performances over state-of-the-art methods by a large margin (\ie, 6.1$\%$, 4.6$\%$ and 3.2$\%$ of Mean Recall (mR@100)) on Visual Genome dataset.

\section{Related work}
\label{sec:related work}
\noindent\textbf{Scene Graph Generation}:
Suffering from biased prediction, today's SGG task is far from practical. To deal with the problem, some methods~\cite{tde,mr,bgnn} are proposed to balance discriminating process in accordance with class distribution or visual clues. \cite{cogtree,pcpl} explore predicate correlations with hierarchical or global structures to discriminate predicates. While correlations among predicates varies with contexts, it is neither hierarchical nor global. Thus, we focus on discriminating hard-to-distinguish predicates with pair-wise predicate correlations, constructed as a predicate graph.




\noindent\textbf{Long-Tailed Distribution Classification}:
To solve long-tailed problem, various distribution-based re-balancing learning strategies~\cite{adaptive,eqb,seesaw} have been proposed. However, besides class distribution, correlations are crucial for differentiating hard-to-distinguish predicates in SGG. Therefore, in this work, we take advantage of both predicate distribution and predicate correlations to handle this issue.

\noindent\textbf{Fine-Grained Image Classification}:
Fine-Grained Image Classification aims to recognize hard-to-distinguish objects in a coarse-to-fine manner. Existing methods tackle the problem from two perspectives, representation-encoding~\cite{fg5,fg6} and local recognition~\cite{fg1,fg2}. 
However, due to complex relationships among predicates, such a coarse-to-fine discriminatory manner may fail to differentiate predicates for SGG. Particularly, different predicates may share similar meanings in a specific scenario, while a predicate may have different meanings in different contexts. Instead of hierarchical structures, predicate correlations should be formed as a graph. Concretely, we construct Predicate Lattice to comprehend predicate correlations for predicate discriminating.

\section{Fine-Grained Predicates Learning}

\subsection{Problem Formulation}
Scene graph generation is typically a two-stage multi-class classification task. In the first stage, Faster R-CNN detects instance labels $O=\{o_i\}$, bounding boxes $B=\{b_i\}$, and feature maps $X=\{x_i\}$ within an input image $I$. In the second stage, scene graph models infer the predicate category from subject $i$ to subject $j$, \ie, $R=\{r_{ij}\}$, based on the detection results, \ie, $Pr(R|O,B,X)$.

\begin{figure*}[t]
\centering
\includegraphics[width=0.75\textwidth]{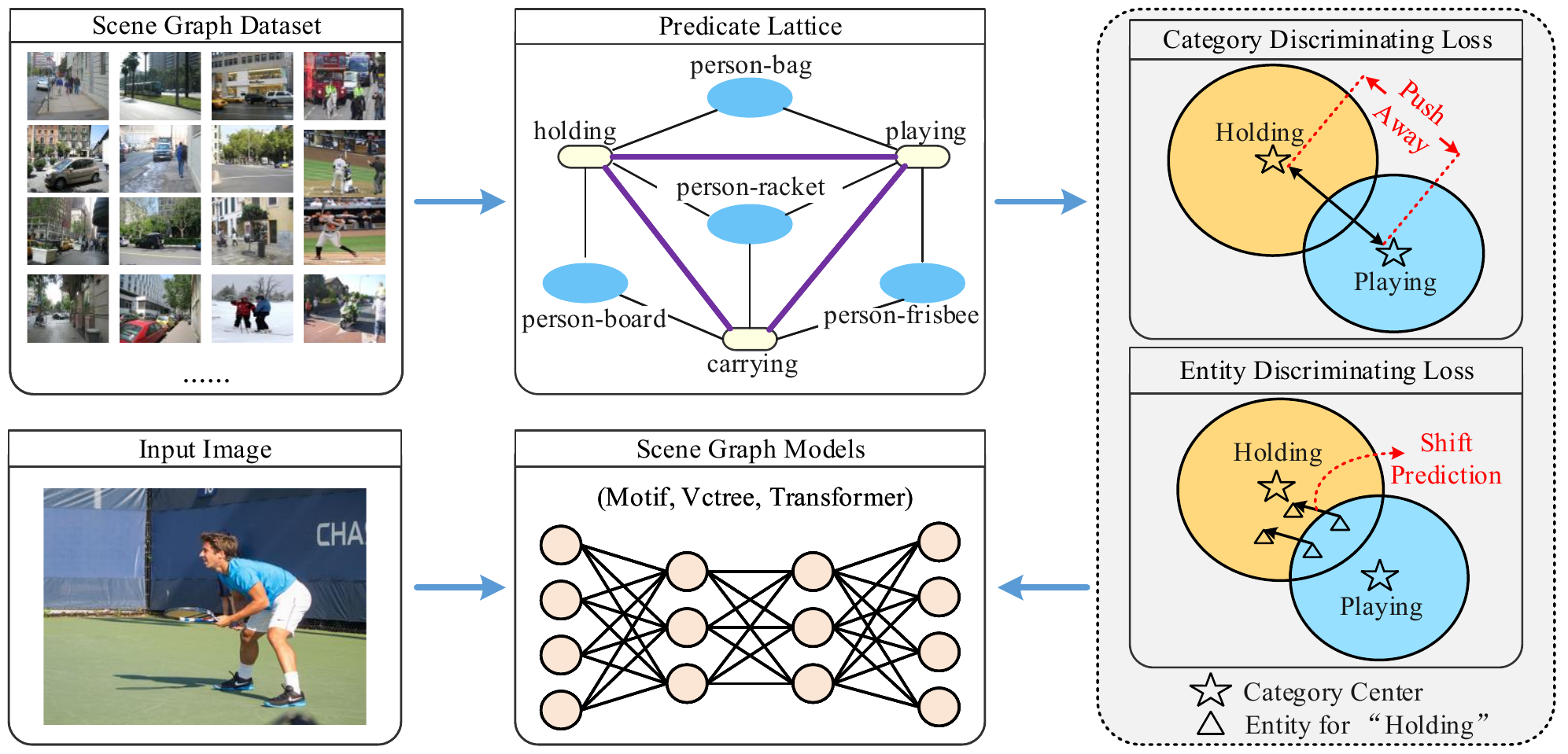} 
\vspace{-0.5em}
\caption{\textbf{The Overview of our Fine-Grained Predicates Learning (FGPL) framework.} It includes three parts: Predicate Lattice, Category Discriminating Loss, Entity Discriminating Loss. Fine-Grained Predicates Learning is incorporated into several state-of-the-art SGG models. Predicate Lattice is constructed from the SGG dataset (Visual Genome) to help understand predicates correlations. With Predicate Lattice, the architecture is optimized with two terms: Category Discriminating Loss and Entity Discriminating Loss.}
\vspace{-1.0em}
\label{fig:framework}
\end{figure*}

Within our Fine-Grained Predicates Learning (FGPL) framework, shown in Fig.~\ref{fig:framework}, we first construct a Predicate Lattice concerning context information to understand ubiquitous correlations among predicates. Then, utilizing the Predicate Lattice, we develop a Category Discriminating Loss and an Entity Discriminating Loss which help SGG models differentiate hard-to-distinguish predicates.

\subsection{Predicate Lattice Construction}
\label{sec:construction}
\begin{figure}[t]
\centering
\includegraphics[width=0.45\textwidth]{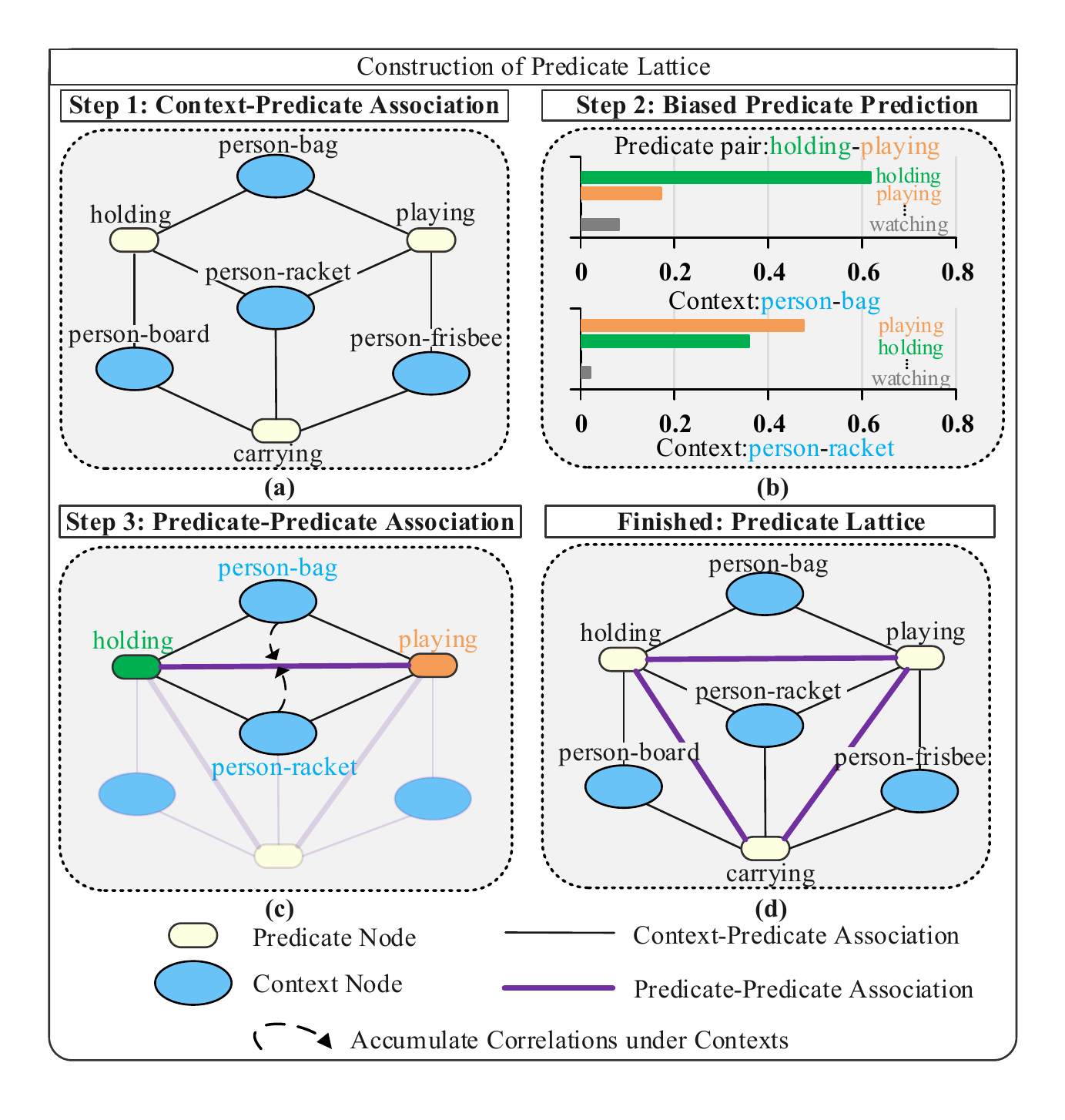} 
	\vspace{-0.5em}
\caption{\textbf{Construction of Predicate Lattice.} The whole process is divided into three steps: (1) Context-Predicate Association; (2) Biased Predicate Prediction; (3) Predicate-Predicate Association.}
\vspace{-1.8em}
\label{fig:lattice}
\end{figure}

To fully understand relationships among predicates, we build a Predicate Lattice, which includes correlations for each pair of predicates concerning contextual information. In general, predicate correlations are acquired under different contexts, since contexts (i.e., visual or semantic information of predicates’ subjects and objects) determine relationships among predicates. Specifically, we extract their contextual-based correlations from biased predictions containing all possible contexts between each pair of predicates. The construction procedure is shown in Fig.~\ref{fig:lattice}.

\noindent\textbf{Context-Predicate Association}: 
We first establish Context-Predicate associations between predicate nodes and context nodes. As contexts determine correlations among predicates, predicate correlations are constructed as a Predicate Lattice containing predicates and related contexts (\ie, visual or semantic information of predicates' subjects and objects). In Fig.~\ref{fig:lattice}(a), we show structures of our Predicate Lattice. There are two kinds of nodes in Predicate Lattice, namely Predicate nodes and Context nodes, which indicate predicate categories and labels of subject-object pairs, respectively. Several predicate nodes connect to the same context node, which denotes that several predicates can describe relationships in the same context. For instance in Fig.~\ref{fig:lattice}(a), both ``holding'' and ``carrying'' can be utilized to describe relationships for ``person-racket''. Specifically, we adopt Frequency model~\cite{motif} to derive every subject-object pair as the context for each predicate from the SGG dataset (VG). Moreover, weights of edges between predicate nodes and context nodes, \ie, $Pr(r_{ij}|o_i,o_j)$, denote occurrence frequency for each ``subject($o_i$)-predicate($r_{ij}$)-object($o_j$)'' triplet in dataset. In this way, we establish connections between predicate nodes and context nodes in Predicate Lattice.

\noindent\textbf{Biased Predicate Prediction}:
To associate predicate pairs with predicate correlations in the next step, we acquire Biased Predicate Prediction from SGG models. Firstly, we incorporate Context-Predicate Association, constructed in step one, into SGG models. 
Particularly, we extract the Context-Predicate Association for each ``subject-predicate-object'' triplet as semantic information. Then, to acquire complete contextual information, we combine semantic information with visual features, \ie, $B={b_i}$ and $X$, of subjects $o_i$ and objects $o_j$ to predict predicates $Pr(r_{ij}|o_i,o_j,b_i,b_j,x_i,x_j)$. With the contextual information, we derive the Biased Predicate Prediction of pre-trained SGG models by inferring on the training set of the SGG dataset concerning all scenarios. In this way, the Biased Predicate Prediction contains predicate predictions under all possible scenarios for each predicate pair. For instance, as shown in Fig.~\ref{fig:lattice}(b), we infer the pre-trained SGG model under all possible scenarios for predicate ``playing'' or ``holding'', such as ``person-racket'' and ``person-bag''.

\noindent\textbf{Predicate-Predicate Association}: At last, we establish Predicate-Predicate Association among predicates with context-based correlations obtained from the Biased Predicate Prediction. 
The Biased Predicate Prediction implies the context-based correlations between each pair of predicates.
For instance, if most samples are predicted as $j$ but labeled as $i$ in ground truth, predicate $i$ is correlated to predicate $j$ in most contexts.
Based on the above observation, we accumulate prediction results from each possible context to obtain the holistic predicate correlations between each pair of predicates, shown in Fig.~\ref{fig:lattice}(c). 
For instance, given predicate pair ``playing-holding'', we gather their correlations under all contexts/scenarios, such as ``person-racket'' and ``person-bag''. Moreover, if predicate $i$ is correlated to predicate $j$ in most contexts, they are prone to be strongly correlated. Therefore, we normalize the gathered predicate correlations as $S=\{s_{ij}\}$ with $s_{ij}\in[0,1]$, which indicates the proportion of samples labeled as $i$ but predicted as $j$. In particular, higher $s_{ij}$ means a stronger correlation between predicate pair $i$ and $j$. Then, we associate predicate pairs with predicate correlations $s_{ij}$. Finally, predicate correlations are formed as a Predicate Lattice, shown in Fig.~\ref{fig:lattice}(d).

\subsection{Category Discriminating Loss}
In this section, we first analyze the limitations of re-weighting methods. Then, we introduce our Category Discriminating Loss (CDL) in detail. 

\noindent\textbf{Limitations of Re-weighting Methods:}
Overall, recent re-weighting methods re-balance the learning process by strengthening the penalty to head classes while scaling down the overwhelming punishment to tail classes. To be specific, the state-of-the-art re-weighting method~\cite{seesaw} adjusts weights for each class in Cross-Entropy Loss on the basis of the proportion of training samples as follows:
	\vspace{-1.0em}
\begin{equation}
\label{balanced_softmax}
	\begin{aligned}
    &\mathcal{L}_{CD}(\eta) = -{\textstyle \sum_{i=1}^{C}}y_ilog(\hat{\phi}_i)~,\\
    &\hat{\phi}_i = \frac{e^{\eta_i}}{ {\textstyle \sum_{j=1}^{C}w_{ij}e^{\eta_j}}}~,
    w_{ij}=\begin{cases}
     {(\frac{{{n_j}}}{{{n_i}}})^\alpha },  & \text{ if } n_j > n_i \;  \\
    1, & \text{ if } n_j \le n_i \; \\
    \end{cases}  ~, \\
	\end{aligned}
\end{equation}
where $\eta=[\eta_1,\eta_2,...,\eta_C]$ and $\hat\phi=[\hat\phi_1,\hat\phi_2,...,\hat\phi_C]$ denote predicted logits and re-weighted probabilities for each class. The label $Y=[y_1,y_2,...,y_C]$ is a one-hot vector. Additionally, $w_{ij}$ denotes the re-weighting factor concerning distribution between positive class $i$ and negative class $j$. Explicitly, $w_{ij}$ is calculated based on the proportion of distribution between class $i$ and $j$, as shown in Eq.~\ref{balanced_softmax}, where $\alpha > 0$. 
\begin{equation}
	\begin{aligned}
\frac{\partial \mathcal{L}_{CD}(\eta)}{\partial \eta_j} = \frac{w_{ij}e^{\eta_j }}{ {\textstyle \sum_{k=1}^{C}w_{ik}e^{\eta_k}}}~.
	\end{aligned}
\label{eq.grad}
\end{equation}
Eq~.\ref{eq.grad} shows negative gradients for category $j$. If positive category $i$ is less frequent than negative category $j$, \ie, $n_j > n_i$ with $w_{ij}>1$, it will strengthen the punishment to negative class $j$. On the contrary, if $n_j \le n_i$ with $w_{ij}=1$, it will degrade the penalty to negative class $j$. Finally, it results in a balanced learning process.


Without considering predicate correlations, re-weighting methods cannot adaptively adjust discriminating process in accordance with difficulty of discrimination, resulting in an inefficient learning process. 
As an inherent characteristic of predicates, predicate correlation reveals difficulty of discrimination for different pairs of predicates. 
However, ignoring predicate correlations in learning process, the re-weighting method roughly reduces negative gradients for all negative predicates with fewer samples than the positive predicate.
As a process to push away the decision boundary from head classes to tail classes, such discriminating process is prone to over-suppress weakly correlated predicate pairs and degrades the learned discriminatory ability of recognizable predicates as maintained in \cite{adaptive,eqb}. 
Take an example among ``on/has/\underline{standing on}'', where ``on-\underline{standing on}'' are strongly correlated and ``has-\underline{standing on}'' are weakly correlated. To prevent the tail class ``standing on'' from being over-suppressed, the re-weighting method roughly degrades negative gradients from both ``on'' and ``has''. Although it strengthens discriminatory power between ``on'' and ``standing on'', it is prone to reduce that between ``has'' and ``standing on'' simultaneously. 



\noindent\textbf{Formulation of CDL:}
Based on the above observations, we should both consider the class distribution and predicate correlations to differentiate hard-to-distinguish predicates. Thus, based on the re-weighting method in Eq.~\ref{balanced_softmax}, we devise Category Discriminating Loss (CDL), which adjusts the re-weighting process according to predicate correlations obtained from Predicate Lattice.
Overall, we utilize predicate correlations $s_{ij}$, defined in Sec.~\ref{sec:construction}, as a signal to adjust the degree of re-weighting between predicates $i$ and $j$.
Especially, we mitigate the magnitude of re-weighting for weakly correlated predicates while strengthening that for strongly correlated ones by setting $w_{ij}$, in Eq.~\ref{balanced_softmax}, with different values.
In this way, we maintain gained discriminatory power among recognizable predicates and further enhance that among hard-to-distinguish ones, shown as below:


\vspace{-0.5em}
\begin{equation}
	\begin{aligned}
w_{ij}&=\begin{cases}
 \mu_{ij}^{\beta}~(\ge 1),  & \text{ if } \mu_{ij}\ge 1 \; and  \; \varphi_{ij}  >\xi~  \\
 1, & \text{ if } \mu_{ij}\ge 1 \; and  \; \varphi_{ij}  \le \xi~ \\
 1, & \text{ if } \mu_{ij} < 1 \; and  \; \varphi_{ij}  >\xi~   \\ 
 \mu_{ij}^\alpha~(< 1),  & \text{ if }\mu_{ij} < 1 \; and  \; \varphi_{ij} \le \xi~  \\
\end{cases}  ~, \\
\mu_{ij} &= \frac{n_{j}}{n_{i}}, ~\varphi_{ij} = \frac{s_{ij}}{s_{ii}},
	\end{aligned}
		\label{predicate_correaltion}
\end{equation}
where $\varphi_{ij}$ is calculated by the proportion between $s_{ij}$ and $s_{ii}$, revealing correlations between predicate $i$ and $j$. In addition, $\alpha$ and $\beta$ are hyper-parameters larger than $0$.
For instance, when $n_j \ge n_i$ ($\mu_{ij} \ge 1$), if $\varphi_{ij}>\xi$ of strongly correlated predicate pair $i$ and $j$, $w_{ij}$ is larger than $1$ to strengthen the punishment on negative predicate $j$. 
In contrast, if $\varphi_{ij}\le\xi$ of weakly correlated predicate pair $i$ and $j$, $w_{ij}$ is set as $1$ to mitigate the magnitude of penalty on negative predicate $j$. That is because the excessive punishment is unnecessary for the weakly correlated predicate $j$, which is easy to distinguish from predicate $i$ for models.
When $n_j < n_i$ ($\mu_{ij} < 1$), we set $w_{ij}\le1$ (including $\varphi_{ij} > \xi$ and $\varphi_{ij} \le \xi$) to relieve the over-suppression from head predicate $i$ to tail one $j$. Moreover, if $\varphi_{ij} \le \xi$, we set $w_{ij} = \mu_{ij}^\alpha~(< 1)$ to mitigate the magnitude of the penalty on negative predicate $j$.

\begin{table*}[ht]
\centering
		\resizebox{0.8\textwidth}{!}{
\begin{tabular}{l|ccccccccc}
	\toprule	\toprule
                \multirow{2}{*}{Method}        &   \multicolumn{3}{c}{Predicate Classification (PredCls)} & \multicolumn{3}{c}{Scene Graph Classification (SGCls)} & \multicolumn{3}{c}{Scene Graph Detection (SGDet)}  \\ \cline{2-10}
                               & mR@20         & mR@50        & mR@100        & mR@20         & mR@50         & mR@100         & mR@20        & mR@50       & mR@100        \\ \hline
				BGNN~\cite{bgnn}       & - & 30.4 & 32.9 & -  & 14.3  & 16.5 & - & 10.7 & 12.6 \\ 
				PCPL ~\cite{pcpl}             & -          & 35.2          & 37.8          & -           & 18.6           & 19.6          & -           & 9.5           & 11.7           \\ 
				TDE-VCTree ~\cite{ssg:vctree,tde}             & 18.4          & 25.4          & 28.7          & 8.9            & 12.2           & 14.0          & 6.9           & 9.3           & 11.1          \\ 
				CogTree-Motif ~\cite{motif,cogtree}& 20.9          & 26.4          & 29.0          & 12.1            & 14.9           & 16.1          & 7.9           & 10.4           & 11.8          \\ 
				CogTree-VCTree ~\cite{ssg:vctree,cogtree}& 22.0          & 27.6          & 29.7          & 15.4            & 18.8           & 19.9          & 7.8           & 10.4           & 12.1          \\ 
				CogTree-Transformer ~\cite{networks:transformer,cogtree}& 22.9          & 28.4          & 31.0          & 13.0            & 15.7           & 16.7          & 7.9           & 11.1           & 12.7          \\ 
				Reweight*-Motif ~\cite{motif,seesaw}& 18.8                  & 28.1          & 33.7            & 10.7           & 15.6          & 18.3           & 7.2           & 10.5        & 13.2   \\ 
				Reweight*-VCTree~\cite{ssg:vctree,seesaw}& 19.4                  & 29.6          & 35.3            & 13.7           & 19.9          & 23.5           & 7.0           & 10.5        & 13.1   \\ 
				Reweight*-Transformer~\cite{networks:transformer,seesaw}& 19.5                  & 28.6          & 34.4            & 11.9           & 17.2          & 20.7           & 8.1           & 11.5        & 14.9   \\
				\hline
				\textbf{FGPL-Motif}         & \textbf{24.3}         & \textbf{33.0}        & \textbf{37.5}         & \textbf{17.1}         & \textbf{21.3}         & \textbf{22.5}          & \textbf{11.1}         & \textbf{15.4}       & \textbf{18.2}  \\ 
				\textbf{FGPL-VCTree}          & \textbf{30.8}         & \textbf{37.5}        & \textbf{40.2}         & \textbf{21.9}         & \textbf{26.2}         & \textbf{27.6}          & \textbf{11.9}         & \textbf{16.2}       & \textbf{19.1} \\ 
				\textbf{FGPL-Transformer}     & \textbf{27.5}         & \textbf{36.4}        & \textbf{40.3}         & \textbf{19.2}         & \textbf{22.6}         & \textbf{24.0}          & \textbf{13.2}         & \textbf{17.4}       & \textbf{20.3}   \\ 
\bottomrule\bottomrule
\end{tabular}}
\vspace{-0.5em}
\caption{\textbf{Comparison between existing methods and FGPL}. * denotes state-of-the-art re-weighting method proposed in~\cite{seesaw}.}

\label{tab.compare}
\vspace{-1.5em}
\end{table*}
\subsection{Entity Discriminating Loss}
Although CDL can effectively differentiate hard-to-distinguish predicates, it still has a limitation: weights assigned to predicates are stable during training, which can neither adapt to the gradually obtained discriminatory power during training nor contexts varied with training samples. Hence, we individually treat prediction results of each sample as signals to adjust the decision boundary. Based on the observations, we propose Entity Discriminating Loss (EDL), which adapts the discriminating process to the learning status and contexts, shown as below:
\begin{equation}
	{\mathcal{L}_{ED}(\eta)} = {\frac{1}{{\left| {\mathcal{V}_i  } \right|}}\sum\limits_{j \in \mathcal{V}_i} {\max (0,{\phi_j} - {\phi_i} + \delta )\frac{{{n_j}}}{{{n_i}}}}~, } 
			\label{eq:EDL}
\end{equation}
where $\mathcal{V}_i$ is defined as a set of strongly correlated predicates selected in reference to predicate correlations $s_{ij}$ in Predicate Lattice. For each predicate category $i$, $M$ predicates with the highest $s_{ij}$ in the Predicate Lattice are chosen to construct $\mathcal{V}_i$. Given the input sample $\eta$, $\phi_i$ and $\phi_j$ are the predicted probabilities for predicates $i$ and $j$, and $\phi_j-\phi_i$ implies the learned discriminatory ability between them during training. The $\delta$ is a hyper-parameter, which denotes prediction margins for predicates. Furthermore, EDL is reduced to zero if predicate pairs are distinguishable enough \ie, ${\phi_i} - {\phi_j} \ge \delta$. Moreover, we also adopt the balancing factor $\frac{n_j}{n_i}$ to alleviate imbalanced gradients between classes with fewer or more observations. 

Finally, we combine CDL and EDL as Eq.~\ref{final}, which distinguishes hard-to-distinguish predicates while maintaining the performance between distinguishable ones.	\vspace{-0.5em}
\begin{equation}
	\begin{aligned}
	\mathcal{L}_{}(\eta)= \mathcal{L}_{CD}(\eta)+\lambda \mathcal{L}_{ED}(\eta)~,
	\end{aligned}
	\label{final}
\end{equation}
where $\mathcal{L}_{CD}$ and $\mathcal{L}_{ED}$ denote Category Discriminating Loss and Entity Discriminating Loss. Futhermore, $\lambda$ is a hyper-parameter balancing CDL and EDL.
\section{Experiments}
\label{sec:experiments}
\subsection{Experiment Setting}
\noindent\textbf{Dataset}: Following previous works~\cite{tde,motif,mr}, we adopt widely used Visual Genome split for scene graph generation. Under the setting, the Visual Genome dataset has $150$ object categories and $50$ relationship categories. Then, we divide the dataset into $70\%$ training set, $30\%$ testing set, and $5k$ images from the training set for validation.

\noindent\textbf{Model Configuration}: For our Fine-Grained Predicates Learning (FGPL) is model-agnostic, following recent works~\cite{prior:iccv}, we incorporate it into VCTree~\cite{ssg:vctree}, Motif~\cite{motif}, and Transformer~\cite{networks:transformer} in the SGG benchmark~\cite{ssg:benchmark}.

\noindent\textbf{Evaluation Metrics}: We evaluate our methods on three sub-tasks in scene graph generation, including PredCls, SGCls, and SGDet. Following recent works~\cite{mr,ssg:vctree,bgnn}, we evaluate the performance of prior methods on mR@K and Group Mean Recall, \ie, head, body, and tail. Besides, we introduce DP@K ($\%$) to indicate models' Discriminatory Power among top-k hard-to-distinguish predicates. Generally, DP@K is calculated by averaging the difference between the proportion of samples correctly predicted as $i$ and the proportion of samples misclassified as hard-to-distinguish predicates $j$ ($j \in \mathcal{V}'_i$). Furthermore, $\mathcal{V}'_i$ is defined as a set of top-k hard-to-distinguish predicates for predicate $i$. Especially, to figure out hard-to-distinguish predicates, we collect a normalized confusion matrix $S' \in \mathbb{R}^{C \times C}$ from the model's prediction results, with $s'_{ij}\in[0,1]$, which denotes the degree of confusion between the predicate pair $i$ and $j$. For each predicate category $i$, $k$ predicates with the highest $s'_{ij}$ are chosen to construct $\mathcal{V}'_i$. In a word, a higher score of DP@K means stronger discriminatory power against hard-to-distinguish predicates.

\subsection{Implementation Details}

\noindent\textbf{Detector}: For object detectors, we utilize the pre-trained Faster R-CNN by~\cite{tde} to detect objects in images. Moreover, weights of the object detectors are frozen during training of scene graph generation for all three sub-tasks. 

\noindent\textbf{Scene Graph Generation Model}: Following~\cite{ssg:benchmark}, baselines are trained with Cross-Entropy Loss and SGD optimizer with an initial learning rate of $0.01$, batch size as $16$.

\noindent\textbf{Fine-Grained Predicates Learning}: We incorporate our FGPL into baselines in Model Zoo~\cite{ssg:benchmark} with the same hyper-parameters for CDL and EDL. In particular, we set $\alpha$, $\beta$, and $\xi$ as $1.5$, $2.0$, and $0.9$ for CDL. Additionally, we set the number of hard-to-distinguish predicates (\ie, $\left| {\mathcal{V}_i } \right|$) as $5$ for EDL. Furthermore, the boundary margin $\delta$, and the hyper-parameter $\lambda$ are set as $0.5$ and $0.1$, respectively.

\subsection{Comparison with State of the Arts}

We evaluate our FGPL by incorporating them into three SGG baselines, namely Transformer~\cite{networks:transformer}, Motif~\cite{motif}, and VCTree~\cite{ssg:vctree}. Quantitative results compared with state-of-the-art methods on Visual Genome are shown in Tab.~\ref{tab.compare}. Specifically, FGPL-Motif, FGPL-VCTree, and FGPL-Transformer outperform CogTree-Motif, CogTree-VCTree, and CogTree-Transformer with consistent improvements as $8.5\%$, $10.5\%$, and $9.3\%$ on mR@100 for PredCls, respectively, demonstrating the effectiveness of the Lattice-Structured Predicate Correlation against the Tree-Structured one, \ie, CogTree. It is worth noting that, although Reweight*-Motif, Reweight*-VCTree, and Reweight*-Transformer exceed most of the prior works on all metrics, FGPL-Motif, FGPL-VCTree, and FGPL-Transformer still achieve a large margin of improvements by $3.8\%$, $4.9\%$, and $5.9\%$ on mR@100 for PredCls, verifying the significant efficacy of FGPL for improving discriminatory power over predicates. Intuitively, fully understanding relationships over predicates, our method can adjust the re-weighting process based on predicate correlations, enhancing the discriminatory ability over predicates.

\begin{table*}[ht]
	\centering
		\resizebox{0.9\textwidth}{!}{

\begin{tabular}{l|ccccccccc}
	\toprule	\toprule
                        \multirow{2}{*}{Method}    &         \multicolumn{3}{c}{Predicate Classification (PredCls)} & \multicolumn{3}{c}{Scene Graph Classification (SGCls)} & \multicolumn{3}{c}{Scene Graph Detection (SGDet)}  \\ \cline{2-10}
                  & mR@20         & mR@50        & mR@100        & mR@20         & mR@50         & mR@100         & mR@20        & mR@50       & mR@100        \\ \hline
            Transformer                & 12.4          & 16.0         & 17.5          & 7.7           & 9.6           & 10.2           & 5.3          & 7.3         & 8.8           \\
            Transformer-FGPL(CDL)      & 23.0 $\uparrow$ \textbf{10.6}        & 31.4 $\uparrow$ \textbf{15.4}       & 35.4 $\uparrow$ \textbf{17.9}         & 14.3 $\uparrow$ \textbf{6.6}          & 18.9 $\uparrow$ \textbf{9.3}        & 21.2 $\uparrow$ \textbf{11.0}              & 9.4 $\uparrow$ \textbf{4.1}        & 13.3 $\uparrow$ \textbf{6.0}       & 16.5 $\uparrow$ \textbf{7.7}       \\
            Transformer-FGPL(CDL+EDL)  & 27.5 $\uparrow$ \textbf{15.1} & 36.4 $\uparrow$ \textbf{20.4}       & 40.3 $\uparrow$ \textbf{22.8}        & 19.2 $\uparrow$ \textbf{11.5}        & 22.6 $\uparrow$ \textbf{13.0}        & 24.0 $\uparrow$ \textbf{13.8}              & 13.2 $\uparrow$ \textbf{7.9}        & 17.4 $\uparrow$ \textbf{10.1}        & 20.3 $\uparrow$ \textbf{11.5}          \\ \hline
            VCTree                     & 11.7          & 14.9         & 16.1          & 6.2           & 7.5           & 7.9            & 4.2          & 5.7         & 6.9           \\
            VCTree-FGPL(CDL)           & 23.0 $\uparrow$ \textbf{11.3}         & 31.6  $\uparrow$ \textbf{16.7}       & 35.3 $\uparrow$ \textbf{19.2}         & 15.7 $\uparrow$ \textbf{9.5}          & 21.1 $\uparrow$ \textbf{13.6}         & 23.3 $\uparrow$ \textbf{15.4}   & 11.0 $\uparrow$ \textbf{6.8}       & 14.7 $\uparrow$ \textbf{9.0}       & 17.5 $\uparrow$ \textbf{10.6}        \\
            VCTree-FGPL(CDL+EDL)       & 30.8 $\uparrow$ \textbf{19.1}         & 37.5 $\uparrow$ \textbf{22.6}        & 40.2 $\uparrow$ \textbf{24.1}         & 21.9 $\uparrow$ \textbf{15.7}         & 26.2 $\uparrow$ \textbf{18.7}         & 27.6 $\uparrow$ \textbf{19.7}          & 11.9 $\uparrow$ \textbf{7.7}          & 16.2 $\uparrow$ \textbf{10.5}        & 19.1 $\uparrow$ \textbf{12.2}          \\ \hline
            Motif             & 11.5          & 14.6         & 15.8          & 6.5           & 8.0           & 8.5            & 4.1          & 5.5         & 6.8           \\
            Motif-FGPL(CDL)         & 22.2 $\uparrow$ \textbf{10.7}         & 30.3 $\uparrow$ \textbf{15.7}        & 34.4 $\uparrow$ \textbf{18.6}          & 12.6 $\uparrow$ \textbf{6.1}          & 16.7 $\uparrow$ \textbf{8.7}        & 18.5 $\uparrow$ \textbf{10.0}         & 8.2 $\uparrow$ \textbf{4.1}        & 11.6 $\uparrow$ \textbf{6.1}      & 14.3 $\uparrow$ \textbf{7.5}        \\
            Motif-FGPL(CDL+EDL)    & 24.3 $\uparrow$ \textbf{12.8}        & 33.0 $\uparrow$ \textbf{18.4}         & 37.5 $\uparrow$ \textbf{21.7}          & 17.1 $\uparrow$ \textbf{10.6}           & 21.3 $\uparrow$ \textbf{13.3}          & 22.5 $\uparrow$ \textbf{14.0}           & 11.1 $\uparrow$ \textbf{7.0}        & 15.4 $\uparrow$ \textbf{9.9}       & 18.2 $\uparrow$ \textbf{11.4}        \\
          \bottomrule\bottomrule
\end{tabular}
}\vspace{-0.5em}
	\caption{\textbf{Quantitative results on generalizability of CDL and EDL in FGPL.} We validate generalization capability of our proposed components, \ie, Entity Discriminating Loss (EDL) and Category Discriminating Loss (CDL), in comparison with baselines.}
	\label{tab.generation1}
	\vspace{-1.5em}
\end{table*}

\subsection{Generalization on SGG Models}

To verify that both CDL and EDL of FGPL are plug-and-play, we incorporate them into different benchmark models, including Transformer, VCTree, and Motif. Quantitative results on Visual Genome are shown in Tab.~\ref{tab.generation1}. From Tab.~\ref{tab.generation1}, compared with baselines, we observe considerate improvements on Transformer-FGPL (CDL) ($17.5\%$ \textit{vs.}~$35.4\%$), VCTree-FGPL (CDL) ($16.1\%$ \textit{vs.}~$35.3\%$), Motif-FGPL (CDL) ($15.8\%$ \textit{vs.}~$34.4\%$) on mR@100 of PredCls task, showing notable generalizability for FGPL (CDL). The reason lies in the fact that CDL helps to figure out and differentiate hard-to-distinguish predicates. Furthermore, after being integrated with FGPL (EDL), our Transformer-FGPL (CDL+EDL), VCTree-FGPL (CDL+EDL), and Motif-FGPL (CDL+EDL) achieve further progress as $4.9\%$, $4.9\%$, and $3.1\%$ on mR@100 of PredCls task, which manifests the great compatibility of our FGPL (EDL). The possible reason is that EDL adjusts the learning process according to the discriminatory ability and contexts varied with learning process and training samples, respectively.

\subsection{Predicate Discrimination of FGPL}
\label{sec.sdl}
We observe that FGPL helps SGG models differentiate hard-to-distinguish predicates, and hence give quantitative and qualitative studies to obtain deep insights into FGPL.

\begin{figure}[tbp]
	\centering
	\subfloat{\includegraphics[width=0.45\linewidth]{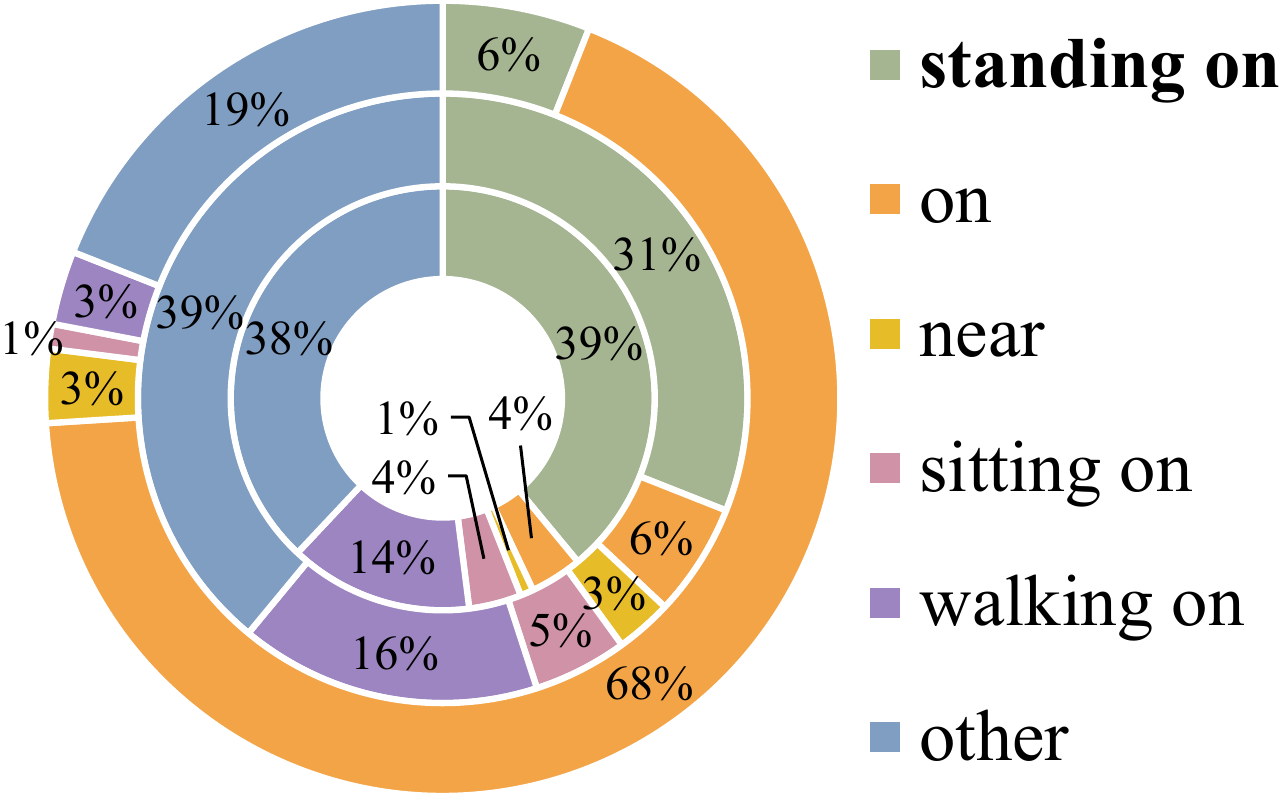}}\quad
	\subfloat{\includegraphics[width=0.45\linewidth]{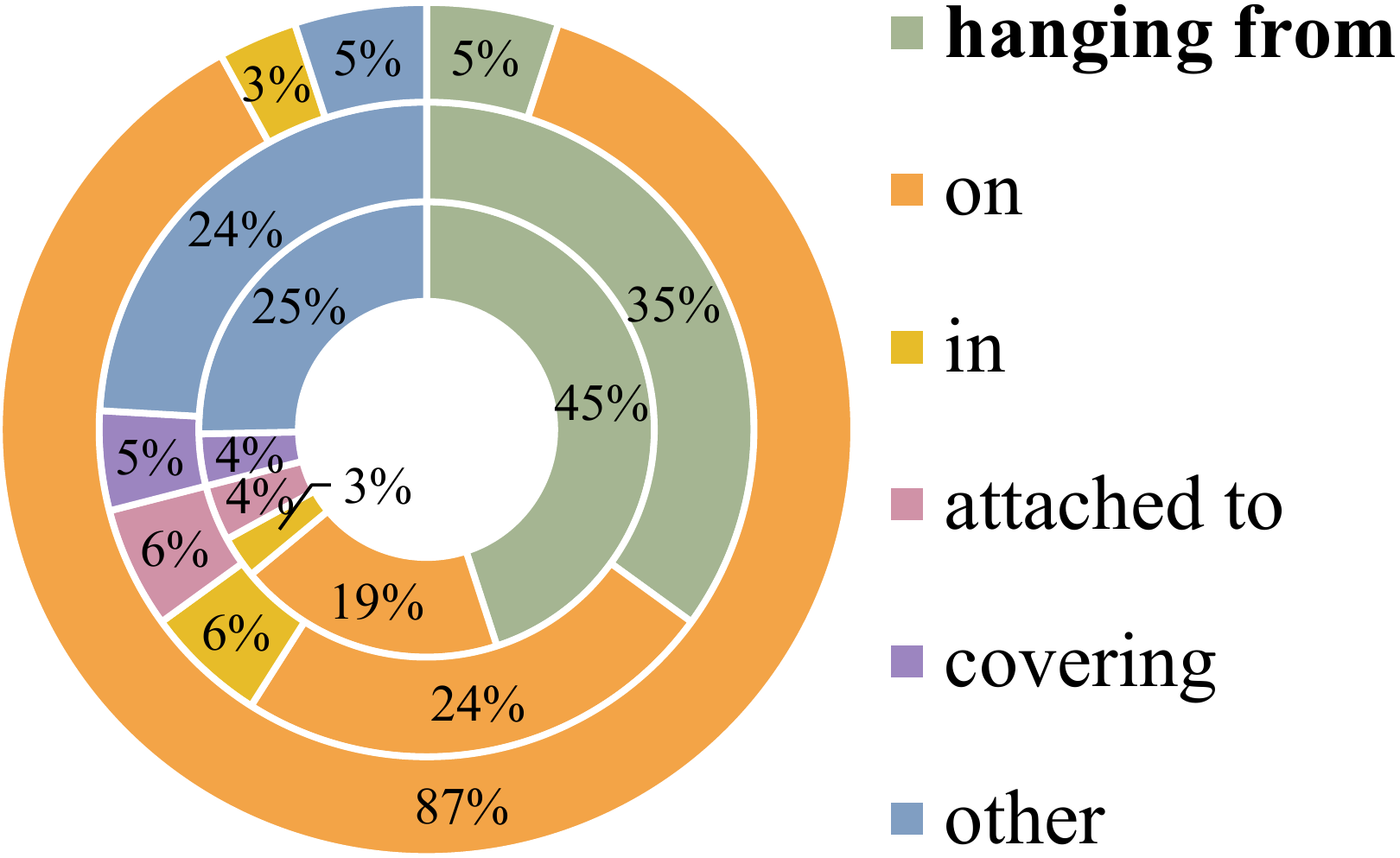}}\\	
	\vspace{-0.5em}
	\caption{\textbf{The effectiveness of FGPL among hard-to-distinguish predicates.} The inner-ring, middle-ring, and outer-ring represent prediction distribution of hard-to-distinguish predicates acquired from Transformer (FGPL), Transformer (Re-weight), and Transformer, respectively, for samples with ground truth as \textbf{``standing on''} on the left, \textbf{``hanging from''} on the right. }
	\vspace{-1.5em}
\label{fig:ring}
\end{figure}

\noindent\textbf{Quantitative Analysis:} As hypothesized, our FGPL improves discriminatory power among hard-to-distinguish predicates while preserving distinguishable ones compared with re-weighting methods. Accordingly, we conduct experiments among three settings to testify our hypothesis: 1) Baselines with traditional Cross-Entropy Loss. 2) Baselines with the state-of-the-art re-weighting method in~\cite{seesaw}. 3) Baselines with our FGPL. To focus on predictions of predicates, we only conduct experiments on PredCls task.
Tab.~\ref{tab.hard-to-distinguishness} presents comparisons among three settings on Transformer, VCTree, and Motif. Besides mR@50, we also evaluate them with DP@K to show discriminatory ability among hard-to-distinguish predicates. After being integrated with FGPL, Transformer (FGPL), VCTree (FGPL), and Motif (FGPL) greatly surpass baselines on DP@10 with a large margin as $22.9\%$ , $22.1\%$, and $22.1\%$. It provides direct evidence that our FGPL considerably improves discriminatory power against hard-to-distinguish predicates. It is also important to note that Transformer (FGPL), VCTree (FGPL), and Motif (FGPL) achieve consistent progress on DP@10 compared with Transformer (Re-weight), VCTree (Re-weight), and Motif (Re-weight). It reflects that our FGPL improves discriminatory ability over the re-weighting method~\cite{seesaw} to generate fine-grained predicates. One possible reason is that FGPL makes the learning process both adapt to correlations of predicates and inherent contextual information of each sample, strengthening discriminatory power against hard-to-distinguish predicates.
\noindent\textbf{Qualitative Analysis:} For an intuitive illustration of FGPL's discriminatory power among hard-to-distinguish predicates, we visualize discrimination among hard-to-distinguish predicates of Transformer, Transformer (Re-weight), and Transformer (FGPL), shown in Fig.~\ref{fig:ring}. The proportion of rings indicates the distribution of prediction results, including hard-to-distinguish predicates $j$ and ground truth predicates $i$, for all samples with ground truth $i$. For predicate ``standing on'' in Fig.~\ref{fig:ring}, Transformer struggles to distinguish it from its correlated predicates, \eg, ``in'' or ``on''. Besides, Transformer (Re-weight) fails to distinct among hard-to-distinguish predicates, \eg, ``standing on'', ``walking on'', and ``sitting on''. For Transformer (FGPL), the proportion of correctly classified samples rises from $6\%$ to $39\%$ compared with Transformer. Meanwhile, hard-to-distinguish predicates are more recognizable than Transformer (Re-weight), \ie, ``walking on'' dropping from $16\%$ to $14\%$ and ``sitting on'' from $5\%$ to $4\%$. Consequently, the results validate our FGPL's efficiency of discriminatory ability against hard-to-distinguish predicates.  

\begin{table}[t]
	\centering
		\resizebox{0.8\linewidth}{!}{
            \begin{tabular}{l|cccc}
	            \toprule	\toprule
            \multirow{2}{*}{Method} & \multicolumn{4}{c}{Predicate Classification (PredCls)}   \\\cline{2-5}
                                      & mR@50        & DP@1       & DP@5         & DP@10 \\ \hline
            Transformer               & 16.0          & 9.9        & 15.6        & 17.4           \\
            Transformer (Re-weight)   & 28.6          & 25.3         & 33.3       & 36.1\\
            Transformer (FGPL)         &\textbf{36.4}         & \textbf{30.1}     &\textbf{37.9}         &\textbf{40.3}        \\\hline
            VCTree                    & 14.9          & 10.5          & 14.1      & 15.7        \\
            VCTree (Re-weight)        & 29.6          & 26.2          & 33.9      & 36.5  \\
            VCTree (FGPL)              &\textbf{37.5}         &\textbf{27.1}       &\textbf{35.4}          & \textbf{37.8}            \\ \hline
            Motif                     & 14.6          & 10.0          & 15.1      & 16.6               \\
            Motif (Re-weight)         & 29.6          & 25.6          & 33.0      & 35.6                  \\
            Motif (FGPL)               & \textbf{33.0}          & \textbf{28.6}         & \textbf{36.1}         & \textbf{38.7}        \\
           \bottomrule\bottomrule
\end{tabular}}
\vspace{-0.5em}
	\caption{\textbf{Quantitative results on discriminatory power} of top-k hard-to-distinguish Predicates (DP@K(\%)) on PredCls. }
	\label{tab.hard-to-distinguishness}
	\vspace{-0.7em}
\end{table}

\subsection{Ablation Study}
To deeply investigate our FGPL, we further study different ablation variants of CDL and EDL on PredCls task.

\noindent\textbf{Entity Discriminating Loss}: To validate the superiority for each component of Entity Discriminating Loss, \ie, Predicate Correlation (PC) and Balancing Factor (BF), we experiment with the following four settings: 1) Transformer with EDL (without PC and BF). 2) Transformer with EDL (without PC), \ie, setting $\mathcal{V}_i$ in Eq.~\ref{eq:EDL} as a set containing all predicate categories. 3) Transformer with EDL (without BF), \ie, removing the balancing factor $\frac{n_{j}}{n_{i}}$ in Eq.~\ref{eq:EDL}. 4) Transformer with EDL (with PC and BF). The experimental results are shown in Tab.~\ref{tab:ablation_EA2}. Without Predicate Correlation (PC), we observe a steep decrease on mR@50 ($22.0\%$ \textit{vs.}~$17.0\%$) and Group Mean Recall (head:$39.2\%$ \textit{vs.}~$37.2\%$, body:$19.7\%$ \textit{vs.}~$11.4\%$, tail:$7.4\%$ \textit{vs.}~$3.7\%$). It verifies the usefulness of PC for improving discriminatory capability for SGG models. The possible reason is that EDL (PC) explores the underlying context information within each entity and adjusts the discriminating process based on the gradually obtained discriminatory capability to alleviate the issue of imbalanced learning. Additionally, it can be observed that trained without BF, there is a substantial reduction on mR@50 ($22.0\%$ \textit{vs.}~$18.9\%$) and Group Mean Recall (head:$39.2\%$ \textit{vs.}~$38.4\%$, body:$19.7\%$ \textit{vs.}~$16.3\%$, tail:$7.4\%$ \textit{vs.}~$5.7\%$), demonstrating the efficacy of BF for a more efficient learning process. We think this may be caused by alleviating over-suppression to tail classes, which leads to a balanced discriminating process among classes with different frequency. At last, when both discarding PC and BF, we observe a larger margin of reduction on mR@50 and Group Mean Recall, demonstrating effectiveness of PC and BF. 
\begin{table}[]
	\centering
	\resizebox{0.8\linewidth}{!}{
	\begin{tabular}{cc|cccc}
	\toprule \toprule
		\multicolumn{2}{c}{EDL}     & \multicolumn{4}{c}{Predicate Classification (PredCls)} \\ \hline
		PC  & BF   & mR@50   & head (16) & body (17) & tail (17) \\ \hline
		$\times$             & $\times$               & 16.2         & 36.5        & 10.5      & 2.5      \\
		$\times$             & $\checkmark$           & 17.0         & 37.2        & 11.4      & 3.7 \\
		$\checkmark$         & $\times$               & 18.9         & 38.4        & 16.3      & 5.7     \\ 
		$\checkmark$         & $\checkmark$           & \textbf{22.0}         & \textbf{39.2}        & \textbf{19.7}      & \textbf{7.4}        \\ \bottomrule
		\bottomrule
	\end{tabular}}
	\vspace{-0.5em}
	\caption{\textbf{Ablation study on each component of EDL}. PC and BF denote Predicate Correlation and Balancing Factor, respectively. The results are obtained with Transformer as the baseline.}
	\label{tab:ablation_EA2}
\end{table}
\noindent\textbf{Category Discriminating Loss}: We explore the effectiveness of the Predicate Correlation (PC) and the Re-weighting Factor (RF) of Category Discriminating Loss. To be specific, we discard PC by ignoring $\varphi_{ij}  >\xi$ and $\varphi_{ij} \le\xi$ in Eq.~\ref{predicate_correaltion}. Besides, we discard RF by setting Re-weighting Factor $w_{ij}$ as 1 for all predicate pairs $i$  and $j$ in Eq.~\ref{balanced_softmax}. The results are shown in Tab.~\ref{tab:ablationCA_1}. It is worth noting that CDL (RF) leads to notable progress on mR@50 and Group Mean Recall, which proves the efficacy of RF for keeping a balanced learning process. Furthermore, CDL outperforms the baseline with a considerable margin after being integrated with PC. We believe that adjusting the re-weighting process according to PC, CDL improves the discriminatory power among hard-to-distinguish predicates while maintaining the original discriminating ability among recognizable ones.

%

\begin{table}[]
	\centering
	\resizebox{0.8\linewidth}{!}{
	\begin{tabular}{cc|cccc}
	\toprule	\toprule
		\multicolumn{2}{c}{CDL}                       & \multicolumn{4}{c}{Predicate Classification (PredCls)} \\ \hline
		PC              & RF                             & mR@50        & head (16)        & body (17)      & tail (17)\\ \hline
		$\times$        & $\times$                       & 16.0         & 36.6        & 10.1      & 2.3\\
		$\times$        & $\checkmark$                   & 28.6         & 32.5        & 30.4      & 22.9\\
		$\checkmark$    & $\checkmark$                   & \textbf{31.4}         & \textbf{37.7}        & \textbf{33.5}      & \textbf{23.3}      \\  \bottomrule\bottomrule
	\end{tabular}}
		\vspace{-0.5em}
	\caption{\textbf{Ablation study on PC and RF of CDL}. PC denotes Predicate Correlation. RF denotes the Re-weighting Factor. The results are obtained with Transformer as the baseline.}
	\label{tab:ablationCA_1}
	\vspace{-1.5em}
\end{table}

\subsection{Visualization Results}

\begin{figure}[t]
\begin{center}
\includegraphics[width=0.46\textwidth]{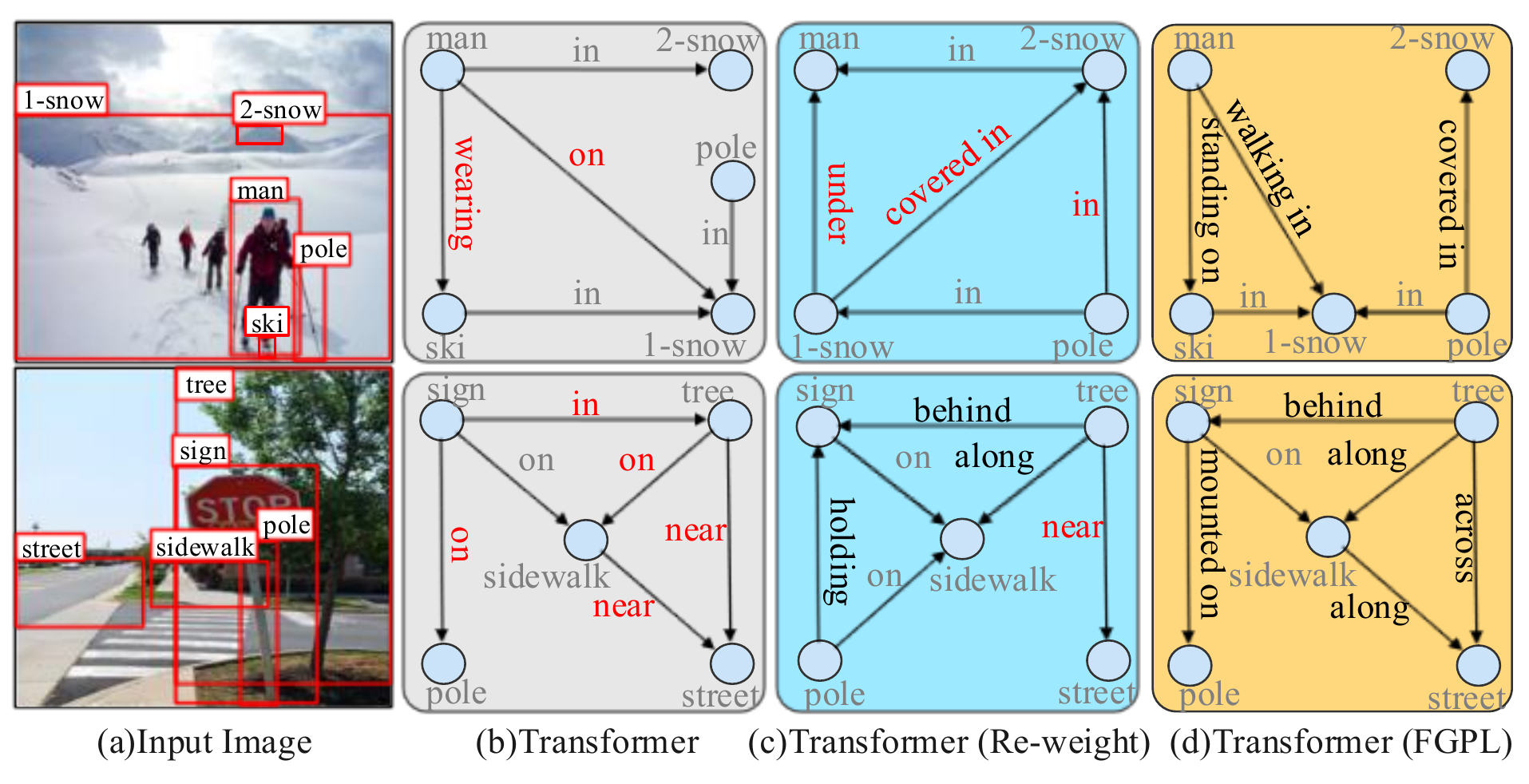}
\vspace{-1.0em}
\caption{\textbf{Visualization results} of Transformer, Transformer (Re-weight), and Transformer (FGPL) on PredCls.}
\label{fig:visual}
\label{visual}
\end{center}
\vspace{-2.5em}
\end{figure}

To intuitively illustrate the effectiveness of our proposed FGPL, we make comparisons among scene graphs generated by Transformer, Transformer (Re-weight), and Transformer (FGPL) with the same input images in Fig.~\ref{visual}. We observe that Transformer (FPL) is capable of generating more fine-grained relationships between objects than Transformer and Transformer (Re-weight), such as ``man-walking in-snow'' rather than ``man-on-snow'', ``tree-across-street'' instead of ``tree-near-street'', and ``sidewalk-along-street'' as opposed ``sidewalk-near-street''.

\section{Conclusion}

\label{sec:conclusions}
In this work, we propose a plug-and-play Fine-Grained Predicates Learning (FGPL) framework for scene graph generation. We devise a Predicate Lattice to help understand predicates correlation concerning all scenarios in the SGG dataset. Based on the Predicate Lattice, we develop a Category Discriminating Loss (CDL) and an Entity Discriminating Loss (EDL), which help differentiate hard-to-distinguish predicates while maintaining learned discriminatory power over recognizable ones. Experiments show that our FGPL can differentiate hard-to-distinguish predicates. When being integrated with our FGPL, several benchmark models achieve superior performance than existing methods, showing the great generability of our FGPL. 
\\ \hspace*{\fill} \\
\noindent\textbf{Broader Impact.} Our research helps reduce the cost of collecting annotations for real-world scenes in applications of scenario understanding. The positive effect of our method on society is in making scenario understanding more efficient for organizations and people. The negative effect of our method on society is that in a deep learning manner, our method is susceptible to adversarial attacks. Therefore, there is a challenge to make people get misunderstood with tampered scenario information. Since training samples come from real-world scenes, it may cause the invasion of personal privacy. Thus, the dataset should be carefully used concerning copyrights and privacy problems. Moreover, the model should be distributed with limitations and regularization in specific scenes. For researchers, we should obey the ethical rules to avoid ethical risks. 
\\ \hspace*{\fill} \\
\noindent\textbf{Acknowledgement.}  This work is supported by National Key Research and Development Program of China (No. 2018AAA0102200), the National Natural Science Foundation of China (Grant No. 62122018, No. 61772116, No. 61872064), Sichuan Science and Technology Program (Grant No.2019JDTD0005).

{

\small
\bibliographystyle{ieee_fullname}
\balance
\bibliography{egbib}

\begin{thebibliography}{10}\itemsep=-1pt

\bibitem{mr}
Tianshui Chen, Weihao Yu, Riquan Chen, and Liang Lin.
\newblock Knowledge-embedded routing network for scene graph generation.
\newblock In {\em CVPR}, 2019.

\bibitem{sample}
Alakh Desai, Tz-Ying Wu, Subarna Tripathi, and Nuno Vasconcelos.
\newblock Learning of visual relations: The devil is in the tails.
\newblock In {\em ICCV}, 2021.

\bibitem{fg1}
Yao Ding, Yanzhao Zhou, Yi Zhu, Qixiang Ye, and Jianbin Jiao.
\newblock Selective sparse sampling for fine-grained image recognition.
\newblock In {\em ICCV}, 2019.

\bibitem{eqb}
Chengjian Feng, Yujie Zhong, and Weilin Huang.
\newblock Exploring classification equilibrium in long-tailed object detection.
\newblock In {\em ICCV}, 2021.

\bibitem{fg5}
Weifeng Ge, Xiangru Lin, and Yizhou Yu.
\newblock Weakly supervised complementary parts models for fine-grained image
  classification from the bottom up.
\newblock In {\em CVPR}, 2019.

\bibitem{cap1}
Jiuxiang Gu, Shafiq~R. Joty, Jianfei Cai, Handong Zhao, Xu Yang, and Gang Wang.
\newblock Unpaired image captioning via scene graph alignments.
\newblock In {\em ICCV}, 2019.

\bibitem{prior:iccv}
Yuyu Guo et~al.
\newblock From general to specific: Informative scene graph generation via
  balance adjustment.
\newblock In {\em ICCV}, 2021.

\bibitem{DBLP:conf/mm/GuoSGS20}
Yuyu Guo, Jingkuan Song, Lianli Gao, and Heng~Tao Shen.
\newblock One-shot scene graph generation.
\newblock In {\em ACMMM}, 2020.

\bibitem{qa4}
Marcel Hildebrandt, Hang Li, Rajat Koner, Volker Tresp, and Stephan
  G{\"{u}}nnemann.
\newblock Scene graph reasoning for visual question answering.
\newblock {\em CoRR}, 2020.

\bibitem{qa1}
Drew~A. Hudson and Christopher~D. Manning.
\newblock {GQA:} {A} new dataset for real-world visual reasoning and
  compositional question answering.
\newblock In {\em CVPR}, 2019.

\bibitem{bgnn}
Rongjie Li et~al.
\newblock Bipartite graph network with adaptive message passing for unbiased
  scene graph generation.
\newblock In {\em CVPR}, 2021.

\bibitem{DBLP:conf/aaai/LiSGLH0G19}
Xiangpeng Li, Jingkuan Song, Lianli Gao, Xianglong Liu, Wenbing Huang, Xiangnan
  He, and Chuang Gan.
\newblock Beyond rnns: Positional self-attention with co-attention for video
  question answering.
\newblock In {\em AAAI}, 2019.

\bibitem{vrr}
Yuanzhi Liang, Yalong Bai, Wei Zhang, Xueming Qian, Li Zhu, and Tao Mei.
\newblock Vrr-vg: Refocusing visually-relevant relationships.
\newblock In {\em ICCV}, 2019.

\bibitem{gps}
Xin Lin, Changxing Ding, Jinquan Zeng, and Dacheng Tao.
\newblock Gps-net: Graph property sensing network for scene graph generation.
\newblock In {\em CVPR}, 2020.

\bibitem{nlp}
L.~John Old.
\newblock An analysis of semantic overlap among english prepositions in roget's
  thesaurus.
\newblock In {\em ACL-SIGSEM}, 2003.

\bibitem{image_retrival}
Brigit Schroeder and Subarna Tripathi.
\newblock Structured query-based image retrieval using scene graphs.
\newblock In {\em CVPR}, 2020.

\bibitem{qa3}
Jiaxin Shi, Hanwang Zhang, and Juanzi Li.
\newblock Explainable and explicit visual reasoning over scene graphs.
\newblock In {\em CVPR}, 2019.

\bibitem{DBLP:conf/ijcai/SongZGS18}
Jingkuan Song, Pengpeng Zeng, Lianli Gao, and Heng~Tao Shen.
\newblock From pixels to objects: Cubic visual attention for visual question
  answering.
\newblock In {\em IJCAI}, 2018.

\bibitem{energy}
Mohammed Suhail, Abhay Mittal, Behjat Siddiquie, Chris Broaddus, Jayan Eledath,
  G{\'{e}}rard~G. Medioni, and Leonid Sigal.
\newblock Energy-based learning for scene graph generation.
\newblock In {\em CVPR}, 2021.

\bibitem{ssg:benchmark}
Kaihua Tang.
\newblock A scene graph generation codebase in pytorch, 2020.
\newblock \url{https://github.com/KaihuaTang/Scene-Graph-Benchmark.pytorch}.

\bibitem{tde}
Kaihua Tang, Yulei Niu, Jianqiang Huang, Jiaxin Shi, and Hanwang Zhang.
\newblock Unbiased scene graph generation from biased training.
\newblock In {\em CVPR}, 2020.

\bibitem{ssg:vctree}
Kaihua Tang, Hanwang Zhang, Baoyuan Wu, Wenhan Luo, and Wei Liu.
\newblock Learning to compose dynamic tree structures for visual contexts.
\newblock In {\em CVPR}, 2019.

\bibitem{networks:transformer}
Ashish Vaswani, Noam Shazeer, Niki Parmar, Jakob Uszkoreit, Llion Jones,
  Aidan~N. Gomez, Lukasz Kaiser, and Illia Polosukhin.
\newblock Attention is all you need.
\newblock In {\em NeurIPS}, 2017.

\bibitem{seesaw}
Jiaqi Wang, Wenwei Zhang, Yuhang Zang, Yuhang Cao, Jiangmiao Pang, Tao Gong,
  Kai Chen, Ziwei Liu, Chen~Change Loy, and Dahua Lin.
\newblock Seesaw loss for long-tailed instance segmentation.
\newblock In {\em CVPR}, 2021.

\bibitem{adaptive}
Tong Wang, Yousong Zhu, Chaoyang Zhao, Wei Zeng, Jinqiao Wang, and Ming Tang.
\newblock Adaptive class suppression loss for long-tail object detection.
\newblock In {\em CVPR}, 2021.

\bibitem{wu2021star}
Bo Wu, Shoubin Yu, Zhenfang Chen, Joshua~B Tenenbaum, and Chuang Gan.
\newblock Star: A benchmark for situated reasoning in real-world videos.
\newblock In {\em NeurIPS}, 2021.

\bibitem{imp}
Danfei Xu, Yuke Zhu, Christopher~B. Choy, and Li Fei{-}Fei.
\newblock Scene graph generation by iterative message passing.
\newblock In {\em CVPR}, 2017.

\bibitem{pcpl}
Shaotian Yan, Chen Shen, Zhongming Jin, Jianqiang Huang, Rongxin Jiang, Yaowu
  Chen, and Xian{-}Sheng Hua.
\newblock {PCPL:} predicate-correlation perception learning for unbiased scene
  graph generation.
\newblock In {\em ACMMM}, 2020.

\bibitem{fg2}
Shaokang Yang, Shuai Liu, Cheng Yang, and Changhu Wang.
\newblock Re-rank coarse classification with local region enhanced features for
  fine-grained image recognition.
\newblock {\em CoRR}, 2021.

\bibitem{cap2}
Xu Yang, Kaihua Tang, Hanwang Zhang, and Jianfei Cai.
\newblock Auto-encoding scene graphs for image captioning.
\newblock In {\em CVPR}, 2019.

\bibitem{cogtree}
Jing Yu, Yuan Chai, Yujing Wang, Yue Hu, and Qi Wu.
\newblock Cogtree: Cognition tree loss for unbiased scene graph generation.
\newblock In {\em IJCAI}, 2021.

\bibitem{motif}
Rowan Zellers, Mark Yatskar, Sam Thomson, and Yejin Choi.
\newblock Neural motifs: Scene graph parsing with global context.
\newblock In {\em CVPR}, 2018.

\bibitem{DBLP:conf/mm/ZengGLJS21}
Pengpeng Zeng, Lianli Gao, Xinyu Lyu, Shuaiqi Jing, and Jingkuan Song.
\newblock Conceptual and syntactical cross-modal alignment with cross-level
  consistency for image-text matching.
\newblock In Heng~Tao Shen, Yueting Zhuang, John~R. Smith, Yang Yang, Pablo
  Cesar, Florian Metze, and Balakrishnan Prabhakaran, editors, {\em ACMMM},
  2021.

\bibitem{fg6}
Heliang Zheng, Jianlong Fu, Zheng{-}Jun Zha, and Jiebo Luo.
\newblock Learning deep bilinear transformation for fine-grained image
  representation.
\newblock In {\em NeurIPS}, 2019.

\end{thebibliography}
}

\end{document}